\documentclass{article}
\PassOptionsToPackage{compress}{natbib}

\usepackage[final]{corl_2024} %
\usepackage{graphics} %
\usepackage{graphicx}
\usepackage{amsmath} %
\usepackage{amssymb}  %
\usepackage{xspace}
\usepackage{algorithm} 
\usepackage{subcaption}
\usepackage{algorithmic} 
\usepackage{booktabs}
\usepackage{multirow}
\usepackage{siunitx}
\usepackage{mdframed,lipsum}
\mdfdefinestyle{MyFrame}{%
    linecolor=black,
    outerlinewidth=2pt,
    outertopmargin=4pt,
    innertopmargin=4pt,
    innerbottommargin=4pt,
    innerrightmargin=4pt,
    innerleftmargin=4pt,
        leftmargin = 4pt,
        rightmargin = 4pt
        }

\usepackage{url}
\usepackage[export]{adjustbox}%
\usepackage{caption}
\usepackage{graphicx}

\usepackage[small]{titlesec}
\titlespacing*{\section}{0pt}{4pt plus 2pt minus 2pt}{2pt plus 1pt minus 1pt}
\titlespacing*{\subsection}{0pt}{1pt plus 1pt minus 1pt}{0pt plus 1pt minus 1pt}
\titlespacing*{\paragraph}{0pt}{0pt plus 0pt minus 4pt}{5pt}

\newcommand{\algo}{NOIR\xspace}

\title{NOIR 2.0: Neural Signal Operated \\ Intelligent Robots for Everyday Activities}

\author{
Tasha Kim$^1$, Yingke Wang$^2$, Hanvit Cho$^3$, Alex Hodges$^2$ \\\\
$^1$Institute for Computational and Mathematical Engineering\\
$^2$Department of Computer Science\\
$^3$Department of Mechanical Engineering\\
\{tashakim, yingkewang, hvcho74, alexh555\}@stanford.edu\\\\
Stanford University
}

\begin{document}
\maketitle
\begin{abstract}
Neural Signal Operated Intelligent Robots (\algo) system is a versatile brain-robot interface that allows humans to control robots for daily tasks using their brain signals. This interface utilizes electroencephalography (EEG) to translate human intentions regarding specific objects and desired actions directly into commands that robots can execute. We present \algo 2.0, an enhanced version of \algo. \algo 2.0 includes faster and more accurate brain decoding algorithms, which reduce task completion time by 46\%. \algo 2.0 uses few-shot robot learning algorithms to adapt to individual users and predict their intentions. The new learning algorithms leverage foundation models for more sample-efficient learning and adaptation (15 demos vs. a single demo), significantly reducing overall human time by 65\%.
\end{abstract}

\section{Introduction}
Brain-robot interfaces (BRIs) represent a major milestone in the fields of art, science, and engineering. The Neural Signal Operated Intelligent Robots (\algo) \cite{zhang2023noir}, unveiled in 2023, is a versatile, intelligent BRI system that employs non-invasive electroencephalography (EEG). The system operates on the concept of hierarchical shared autonomy, where humans set high-level objectives, and the robot carries out these objectives through the execution of detailed motor commands. At the time of its introduction, \algo demonstrated its general-purpose nature by being able to handle a variety of tasks (20 everyday activities) and showing broad accessibility, as it requires minimal training for use by the general public. Moreover, \algo is adaptive and intelligent, equipped with a broad set of skills that enable it to autonomously perform low-level actions. Human intentions are conveyed, interpreted, and executed by the robots through \emph{parameterized primitive skills}, such as \texttt{Pick(obj-A)} or \texttt{MoveTo(x,y)}. Additionally, \algo can learn and adapt to human goals throughout the course of their collaboration.

\algo is built on a modular neural signal decoding pipeline. Decoding human intentions (e.g., ``grasp the mug by the handle") from neural signals is highly complex. Therefore, we break down human intentions into three components: \emph{What} (the object to interact with), \emph{How} (the manner of interaction), and \emph{Where} (the specific location of interaction), demonstrating that these elements can be extracted from various neural data types. These signals, once decoded, map naturally to the robot’s parameterized skills and can be effectively communicated to the robots.

However, there is yet much to be improved for \algo. First, the decoding time and effort were considerably high. Performing the tasks (4-15 primitive skills) took 3 to 43 minutes, and 55\%-85\% of the time was spent on the decoding side. Additionally, the decoding accuracy at test time, especially for skill selection, was relatively low (42\% for 4-way classification, 74\% for 2-way classification). By leveraging the recent progress in neural decoding \cite{chin2009multi}, we show that both the decoding time and accuracy can be significantly improved. 

Second, once a human participant has successfully performed the task multiple times, \algo was able to use its retrieval-based few-shot object and skill selection algorithm to predict human intention, which was shown to save decoding time by 60\%. However, the algorithm was based on a pre-trained R3M model \cite{nair2023r3m}, which requires 15 training trajectories to predict human intention successfully. Although this number does not seem high for typical robot learning research, it is considered impractical for BRIs, especially in clinical trials. We show that by leveraging the recent progress in large, pre-trained vision-language models, we can accurately predict human intention with only one trajectory, making \algo 2.0 more effective and practical.

\section{The \algo 2.0 System}
\label{sec:method}
\begin{figure}
    \centering
    \includegraphics[width=0.95\textwidth]{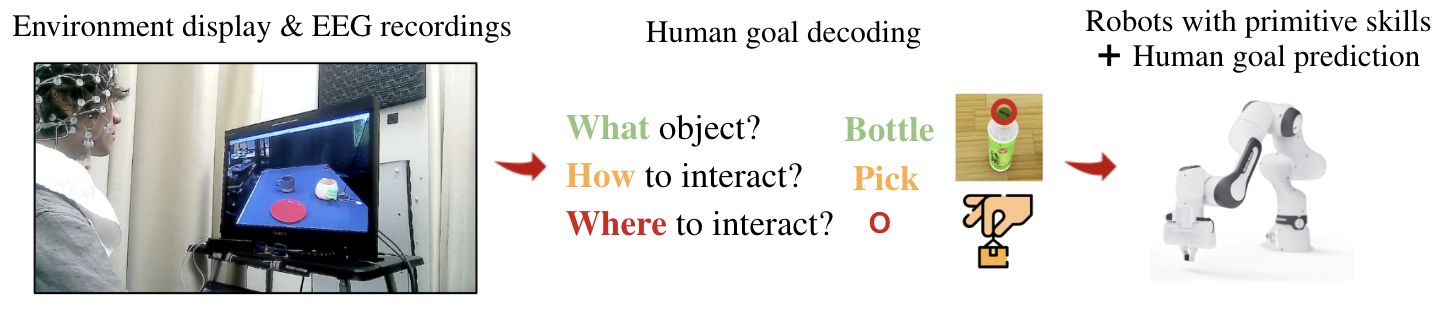}
    \caption{\algo 2.0 system overview. Following \algo, \algo 2.0 implements a modular pipeline for decoding goals from human brain signals, and a robotic system with a library of primitive skills. While minimizing the effort needed for decoding, the robot system is capable of learning to anticipate the goals that humans intend to achieve.}
    \label{fig:method}
\end{figure}

Fig.~\ref{fig:method} shows a schematic representation of the system. In this setup, humans function as planning agents who perceive and communicate behavioral objectives to the robot. The robot, equipped with pre-defined primitive skills then executes these objectives.

\subsection{The Brain: A modular decoding framework}
\label{method:decoding}
We opted for a non-invasive, saline-based EEG system that records the brain's spontaneous electrical activity through electrodes positioned on the scalp. EEG-based BRIs have been utilized in various applications, including prosthetics, wheelchairs, and robots designed for navigation and manipulation~\cite{aljalal2020comprehensive,nicolas2012brain,bi2013eeg,krishnan2016electroencephalography}. Our approach leverages two commonly used EEG signal types in BRIs: steady-state visually evoked potential (SSVEP) and motor imagery (MI).

\begin{figure}
    \centering
    \includegraphics[width=0.95\linewidth]{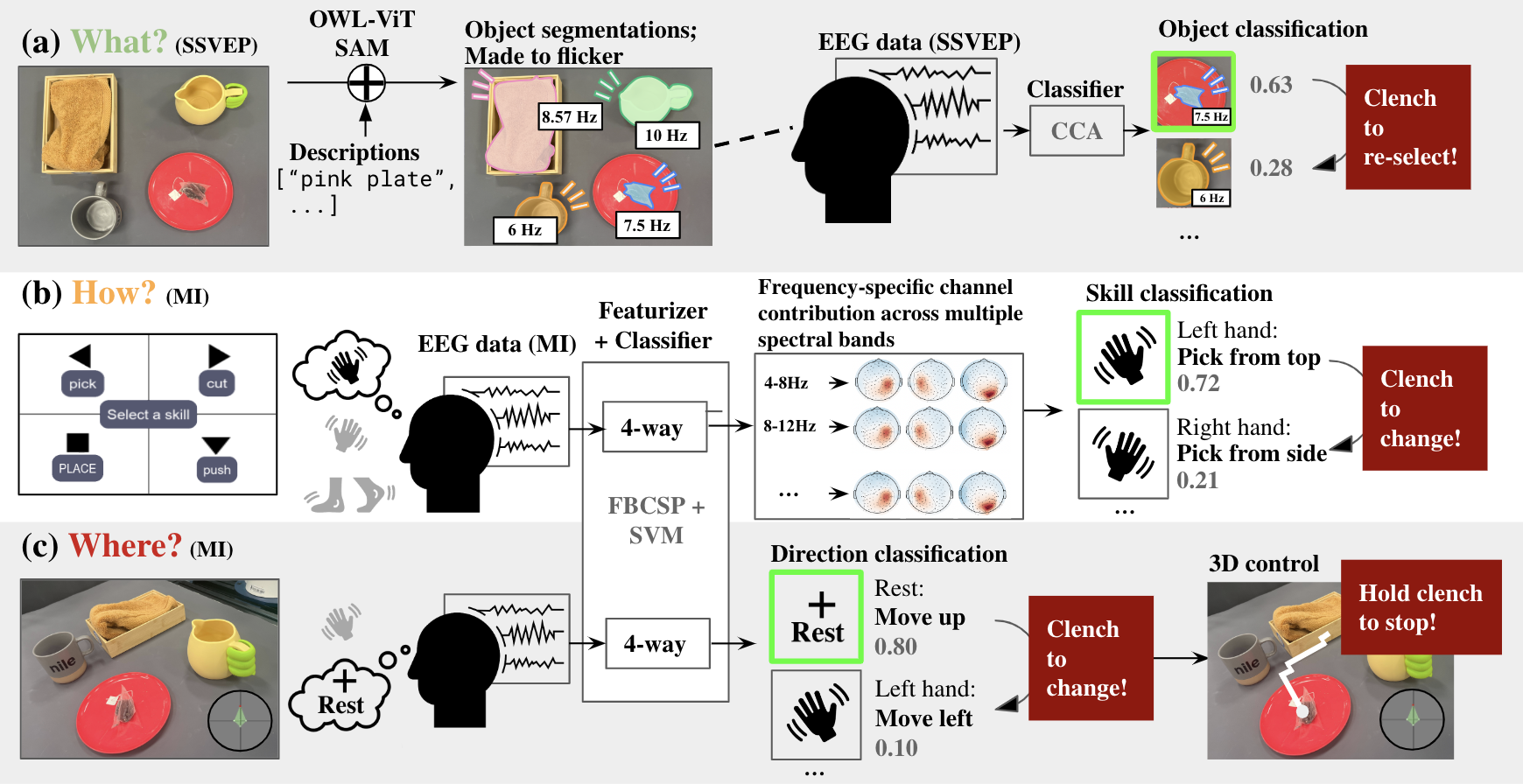}
    \caption{A structured framework for interpreting human objectives from EEG signals that include: (a) \emph{What} object to manipulate, determined by SSVEP signals using CCA classifiers; (b) \emph{How} to engage with the object; and (c) \emph{Where} to interact with the object, deciphered through MI signals using FBCSP+SVM algorithms. A safety mechanism also monitors muscle tension from the jaw to either confirm or reject the decoding results.}
    \label{fig:decoding}
\end{figure}

SSVEP is the brain’s response to external visual stimuli presented at regular intervals \cite{adrian1934berger}. When an individual focuses on a flickering object, the EEG response at the frequency of that stimulus becomes stronger, enabling identification of the object. In contrast, MI is endogenous, requiring the user to mentally simulate specific actions, such as imagining how to manipulate an object. The decoded signals from MI provide insight into how a person intends to interact with the object. As depicted in Fig.~\ref{fig:decoding}, we break down human intention into three parts: (a) \emph{What} object is being manipulated; (b) \emph{How} to engage with the object; and (c) \emph{Where} to engage with it.

\paragraph{Object selection via steady-state visually evoked potential (SSVEP).} 
Once the task setup is displayed on a screen, we first determine the user's target object: each object on the screen is made to flicker at distinct frequencies (Fig.~\ref{fig:decoding}), and when the user focuses on one, it generates an SSVEP response \cite{adrian1934berger}. By detecting which frequency shows a stronger presence in the EEG data, we can identify the flickering visual stimulus and thus which object the user is focusing on. We utilize OWL-ViT \cite{minderer2022simple} for object detection and tracking, which processes images along with object descriptions to generate segmentation masks. By superimposing each mask with the corresponding flicker frequencies ($6Hz,~7.5Hz,8.57Hz$, and $10Hz$\cite{zhu2010survey,kus2013quantification}), we can accurately identify the chosen object when the user concentrates on it for 10 seconds. For this, we only process signals from the visual cortex and apply a notch filter to the data. Next, we employ canonical correlation analysis (CCA) for classification \cite{CCA_SSVEP}. We generate a canonical reference signal (CRS), which consists of $\sin$ and $\cos$ waves corresponding to each frequency and its harmonics. CCA is then used to determine which CRS correlates most strongly with the EEG signal, allowing us to find the object that flickered at the corresponding frequency.

\paragraph{Skills and parameters selection via motor imagery (MI).}
The user selects a skill and its parameters, which we present as a $k$-way ($k \leq 4$) MI classification problem. The goal is to identify which of the $k$ pre-determined actions the user envisions. Unlike SSVEP, brief calibration data (10 min) is required due to the unique MI signal patterns present for each individual. The four categories are: \texttt{Left Hand}, \texttt{Right Hand}, \texttt{Legs}, and \texttt{Rest}, corresponding to the body parts the user imagines moving to perform certain actions (e.g. pressing a pedal with their feet). After displaying the skill options, we capture a 3-second EEG recording and apply a classifier trained on the calibration data to interpret the signals. The user then controls a cursor on the display to the desired skill execution point. To move the cursor continuously along the $xy$ plane, the user is prompted to imagine moving their \texttt{Left} or \texttt{Right Hands} for leftward or rightward movement; \texttt{Legs} or \texttt{Rest} for downward or upward movement, respectively. This method is also used for moving the cursor vertically along the $z$ axis, enabling control in three-dimensional space.

For the decoding process, we specifically utilize EEG channels located near the brain regions associated with motor imagery. The data undergoes band-pass filtering within the $8Hz$ to $30Hz$ range to capture the $\mu$-band and $\beta$-band frequencies, which are known to be lined with MI activity \cite{padfield2019eeg}. The classification algorithm is based on the filter-bank common spatial pattern (FBCSP)~\cite{chin2009multi} algorithm coupled with a support vector machine (SVM) classifier. Due to its simplicity, FBCSP+SVM is explainable and amenable to smaller training datasets. Specifically, the FBCSP algorithm identifies unique spatial patterns across different frequencies, highlighting concentrated activity over the left and right motor regions as well as the visual cortex, the latter being associated with the \texttt{Rest} class. The multi-band nature of FBCSP allows for a more comprehensive analysis of these spatial patterns across different spectral ranges.

\begin{figure}
    \centering
    \includegraphics[width=0.9\linewidth]{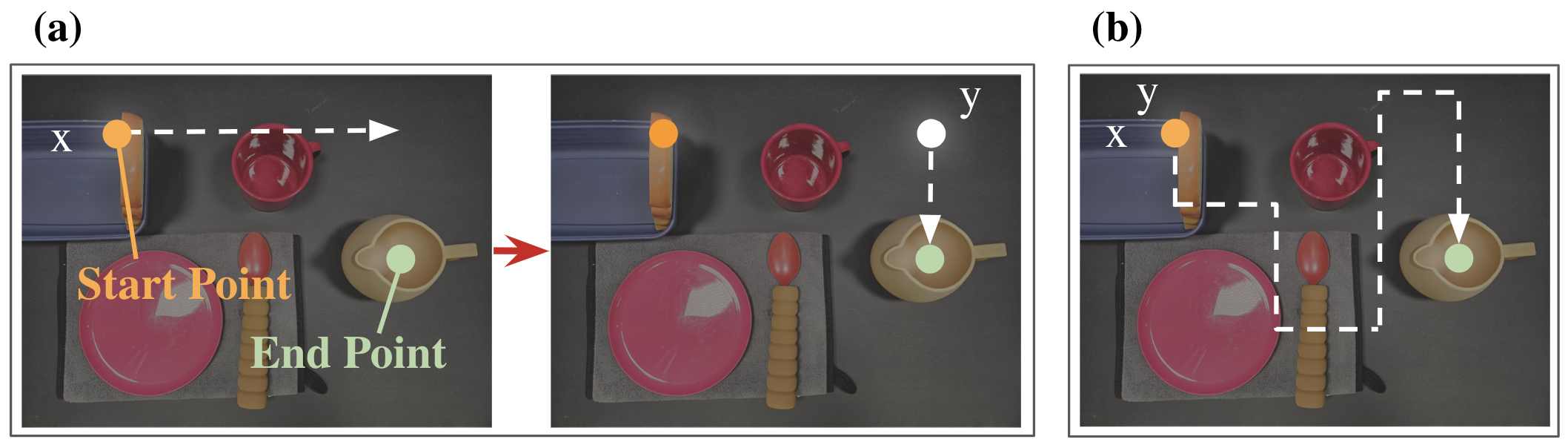}
    \caption{Example cursor trajectories during parameter selection on a graphical user interface: (a) shows how the mouse cursor moves using binary control in NOIR. (b) shows how the mouse cursor moves in all four directions through
    continuous, closed-loop control in NOIR 2.0.
    }
    \label{fig:cursor_movement}
\end{figure}

In Fig.~\ref{fig:cursor_movement}, the evolution of parameter selection methodology from NOIR to NOIR 2.0 is illustrated. In the NOIR system, parameter selection is conducted through binary classification using the CSP algorithm, where the direction is decoded then the cursor first moves along the $x$-axis until the user gives a cue to stop; followed by the $y$-axis then the $z$. However, in NOIR 2.0, the process is refined to enable simultaneous and continuous selection of the $x$ and $y$ directions through 4-way classification. The enhancement leverages MI decoding to facilitate real-time cursor movement, offering a more seamless user experience. Moreover, by introducing a closed-loop feedback mechanism, \algo 2.0 allows users to reject or re-select, and actively adapt to new cursor positions.

\paragraph{Confirmation or interruption through muscle tension.}
As a safety measure, we incorporate a widely used method of capturing electrical signals produced in facial muscle tension (i.e. electromyography, or EMG). These signals are triggered whenever the user frowns, flinches, or clenches their jaw. Since EMG gives strong and highly accurate signals, we use them to confirm or reject selections of objects, skills, or parameters. Using a threshold value established during the calibration phase, \algo 2.0 reliably detects muscle tension using variance-based threshold filters within a short 500-ms window. To enable real-time interruption signaling during human control time compared to \algo, \algo 2.0 filters EEG artifacts that are not generated by the brain such as blinking, lateral eye movement, respiration, and pulse. In addition to this, \algo 2.0 further utilizes spatial filtering, frequency and duration analysis of the signals, and targeted channel selection to more accurately and robustly identify signals to reject.

\subsection{The Robot: Parameterized primitive skills}
\label{method:robot}
Like \algo, \algo 2.0 equips the robots with a collection of parameterized primitive skills~\cite{chitnis2022learning,nasiriany2022augmenting,zhu2021hierarchical,shridhar2022cliport,shridhar2022perceiver,liu2022structformer,xu2021deep,wang2022generalizable,cheng2022guided,agia2022taps,li2023behavior,hiranaka2023primitive}. For our experiments, we use the Franka Emika Panda arm to perform tabletop manipulation tasks. The skills for the Franka robot rely on the operational space pose controller (OSC)~\cite{khatib1987unified}, integrated through the Deoxys API \cite{zhu2023viola}. For instance, the \texttt{Reaching} skill involves generating trajectories via numerical 3D interpolation, based on the robot's current 6D end-effector pose and the target pose. The OSC then directs the robot to follow the trajectory by moving through each waypoint in sequence.

\subsection{Leveraging efficient BRI through robot learning}
\label{method:learning} 
\begin{figure}
    \centering
    \includegraphics[width=0.8\textwidth]{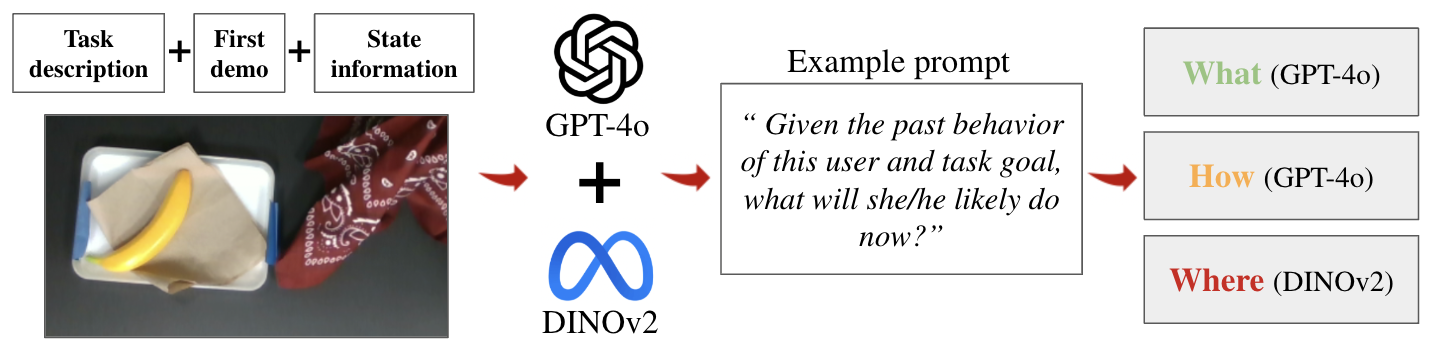}
    \caption{Overview of the robot learning method for \algo 2.0. }
    \label{fig:learning}
\end{figure}

While performing tasks, the robots need to learn the user’s preferences for object, skill, and parameter selections. This allows the robots to predict the user’s intended goals in future trials, making them more autonomous and reducing the need for decoding. It is crucial for the system to learn and generalize effectively since factors like the location, orientation, arrangement, and instances of objects vary across different trials. Additionally, the learning algorithms must be highly sample-efficient, since gathering human data is costly and time-consuming in this setting. The overall design of the learning algorithm is shown in Fig.~\ref{fig:learning}.

\paragraph{Retrieval-based few-shot object and skill selection.}
Human effort can be reduced if the robot intelligently learns to propose the appropriate object-skill selection for a given state in the task. \algo uses retrieval-based imitation learning \cite{mansimov2018simple,nasiriany2023learning,du2023behavior}, which learns a latent state representation (pre-trained R3M model \cite{nair2023r3m} plus a few trainable layers) from observed states. Given a new state observation, it finds the most similar state in the latent space and the corresponding action. However, such a method still requires a handful of demonstration data to be effective (around 15 for \algo), which is impractical in BRI research. With recent progress in large, pre-trained vision language models (VLMs) such as GPT-4o, we show that this can be done with a single training example. 
We design the prompts for the vision-language model (VLM) to assist the following two parts (I) and (II):

\textbf{(I) State understanding and mapping.}

Given a sequence of annotated human demonstration state images, the VLM is tasked with interpreting the spatial relationships between the robotic gripper and objects on the table at each state. It must also comprehend the state of these objects. Putting the task description to use, the model maps each state to relevant objects and their associated skills, while recognizing the logical progression between the states. \textit{Sample prompt:}

\begin{mdframed}
The gripper view image captures what is held in the robot's gripper during the current state. ... The top-view image captures the spatial relationships and status of objects on the table during the current state. ... I will give you a sequence of all the states in the current task, arranged in order. Please do the following:
\begin{enumerate}
    \item Fully understand the object types in the gripper view images and pair them with the `\texttt{obj}' text annotations. ...
    \item Fully understand the objects in the top-view images, the spatial relationships between these objects, and the current status of each object, pairing them with the `\texttt{skill}' text annotations. ...
\end{enumerate}
After that, I will send you two query images. ...
\end{mdframed}

\textbf{(II) State inference and task retrieval.}

Given a query image of the robot's current state, the VLM is expected to infer the robot's current state. It then retrieves the subsequent state by identifying the relevant object and the task to be performed, informed by the prior human demonstration. Finally, it returns the index of the inferred current state. \textit{Sample prompt:}

\begin{mdframed}
Identify the state index from the sequence that best matches this new query pair. Then, return the `\texttt{obj}' and `\texttt{skill}' information of the next state (i.e., \texttt{index + 1}). ... Rank your answer by the order of possibility. ...
\end{mdframed}

\paragraph{One-shot skill parameter learning.}
Selecting parameters requires significant human effort because it involves precise cursor control through continuous MI. \algo 2.0 retains the design of \algo by using a learning algorithm to predict parameters based on an object-skill pair. With DINOv2~\cite{oquab2023dinov2}, we can locate corresponding semantic key points, removing the need for repeatedly specifying parameters. Given a top-view training image ($360\times240$) and a chosen parameter $(x,~y)$, the model predicts the semantically equivalent point in the top-view test images, even when the target object’s position, orientation, instance, or context vary. With a side-view training image ($360\times240$) and a chosen parameter $(z)$, the model first predicts the semantically equivalent point $A$ in the side-view test image, then retrieves the point $B$ on the $z$-axis calculated based on the predicted $(x, ~y)$ that gives the minimal distance from $A$. We convert the pixel coordinates of $B$ into the 3D coordinates that the robot's primitive skill should take in. 

We use a pre-trained DINOv2 model to extract semantic features \cite{oquab2023dinov2}. Both the training and test images are processed by the model to generate 768 patch tokens, forming a pixel-wise feature map of $75\times100$. We then take a $3\times3$ patch centered on the provided training parameter and identify the matching feature in the test image, using cosine similarity as the distance measure. 

\section{Experiments}
 \paragraph{Tasks.} We focus on three tabletop tasks described in \algo: \textsc{WipeSpill}, \textsc{OpenBasket}, and \textsc{PourTea} \cite{zhang2023noir}. For systematic evaluation of the task success, we provide formal definitions of these activities in the BDDL language format \cite{li2023behavior,srivastava2022behavior}, which specify the initial and goal conditions of a task using first-order logic (FOL). 
 
Our human study was approved by the Institutional Review Board (IRB). One healthy participant completed all three tasks above. We employed the EGI NetStation EEG system, which is entirely non-invasive, and makes it suitable for a wide range of participants. Prior to the experiments, the user was introduced to task descriptions and the system interface. During the experiment, the participant remained in a secluded room, stayed still, watched the robot's actions through a screen, and used only their brain signals to interact and communicate with the robot.

\section{Results}

\begin{table}[]
    \centering
    \resizebox{\textwidth}{!}{
    \begin{tabular}{l c c c c c c}
        \toprule
        \multirow{2}{*}{} &
          \multicolumn{3}{c}{\textbf{Time (min.)}} &
          \multicolumn{3}{c}{\textbf{Human Time (min.)}} \\
        \cmidrule(lr){2-4} \cmidrule(lr){5-7}
        \textbf{Task Name} & \textbf{NOIR} & \textbf{NOIR 2.0} & \textbf{NOIR 2.0+Learning} & \textbf{NOIR} & \textbf{NOIR 2.0} & \textbf{NOIR 2.0+Learning} \\ 
        \midrule
        \textsc{WipeSpill}     & 14.74 & 9.12 & 5.46 & 11.65 & 5.12 & 3.15 \\
        \textsc{OpenBasket}    & 15.90 & 6.79 & 5.80 & 13.04 & 2.60 & 1.52 \\
        \textsc{PourTea}       & 13.53 & 8.90 & 12.60 & 11.25 & 6.55 & 7.87 \\
        \midrule
        Avg. Time Reduced (\%) & -- & \textbf{43.82} & \textbf{45.97} & -- & 
        \textbf{60.30} & 
        \textbf{65.11} \\
        \bottomrule
    \end{tabular}
    }
    \vspace{0.05in}
    \caption{\algo vs. \algo 2.0 system performance comparison. Time (min.) indicates the total task completion time for successful trials. Human time (min.) refers to the total time spent by the user, which includes both decision-making and decoding time. Time reduced in both is measured for \algo 2.0 and \algo 2.0 with robot learning (\algo 2.0+Learning) against \algo.}
    \label{tab:system_performance}
\end{table}

\paragraph{System performance.} Table~\ref{tab:system_performance} summarizes the performance based on two metrics: the (a) total time taken to complete a task (Time), and (b) human time spent on the task (Human Time). Note that for all tasks, the task horizon (average number of primitive skills executed) fell in the range 4-6, and the average number of attempts until the first success was 1-2 (1 means success on the first attempt). If the participant encountered a state during task execution from which recovery was not possible, we would reset the environment, and allow them to start the task anew.

For \algo, the average task completion time was $14.72$ minutes across the tasks, and the time humans spent on decision-making and decoding was relatively long ($81.28\%$ of total time). With \algo 2.0, we saw a marked performance improvement in average task completion time at 8.27 minutes, with a 60.30\% decrease in human time spent. Coupled with robot learning, the task completion time was reduced by 45.97\% in total, and human time spent was reduced by 65.11\%.

\paragraph{Decoding accuracy.} 
A key to the success of a BRI system is in the accuracy of decoding brain signals. Tables~\ref{tab:decoding1} and~\ref{tab:decoding2} 
compares \algo vs. \algo 2.0 by summarizing the decoding accuracy at different stages of the pipeline. Our results indicate that using CCA on SSVEP yields an accuracy rate of $88\%$ at task time, which shows that object selection is predominantly precise. The improvement in SSVEP accuracy in \algo 2.0 is attributed to color and contrast optimizations of the masks and background. As for FBCSP+SVM on MI for parameter selection, the $4$-way skill-selection classification models increased the accuracy of skill selection from $42\%$ to $61\%$ at task-time. While the accuracy might not appear high, this is quite competitive given the variations caused by long task duration and differences in setting observed at calibration and task times.

\begin{table}[ht]
    \centering
    \resizebox{\textwidth}{!}{
    \begin{tabular}{lllcc}
        \toprule
         \textbf{Decoding Stage} & \textbf{Signal} & \textbf{Technique} & \textbf{Calibration Acc.} & \textbf{Task-Time Acc.} \\
         \midrule
         Object selection (\emph{What?}) & SSVEP & CCA (4-way) & {--} & 0.81  \\
         Skill selection (\emph{How?}) & MI & CSP+QDA (4-way) & 0.58 & 0.42 \\
         Confirmation / interruption & EMG & Thresholding (2-way) & 1.0 & 1.0  \\
         \bottomrule
    \end{tabular}
    }
    \vspace{0.05in}
    \caption{NOIR decoding accuracy at different stages of the experiment.}
    \label{tab:decoding1}
    
    \vspace{0.3cm} %
    
    \resizebox{\textwidth}{!}{
    \begin{tabular}{lllcc}
        \toprule
         \textbf{Decoding Stage} & \textbf{Signal} & \textbf{Technique} & \textbf{Calibration Acc.} & \textbf{Task-Time Acc.} \\
         \midrule
         Object selection (\emph{What?}) & SSVEP & CCA (4-way) & {--} & \textbf{0.88}  \\
         Skill selection (\emph{How?}) & MI & FBCSP+SVM (4-way) & \textbf{0.64} & \textbf{0.61} \\
         Confirmation / interruption & EMG & Thresholding (2-way) & 1.0 & 1.0  \\
         \bottomrule
    \end{tabular}
    }
    \vspace{0.05in}
    \caption{NOIR 2.0 decoding accuracy at different stages of the experiment.}
    \label{tab:decoding2}
\end{table}

\paragraph{Object and skill selection results.}
We then turn to the question: Does our new robot learning algorithm further improve the efficiency of \algo 2.0? We assess this by evaluating learning in the object and skill selection stage. When an image is presented, a prediction is deemed accurate if and only if it correctly identifies the object and the associated skill. Results are shown in Table~\ref{tab:retrieval}.

\begin{table}[ht]
    \centering
    \begin{tabular}{l l c c}
        \toprule
        \textbf{Model} & \textbf{Task Prediction}  & \textbf{Offline Acc.} & \textbf{Task-Time Acc.} \\
        \midrule
        GPT-4o & Object / skill selection  & 0.94 & 0.83 \\
        DINOv2 & Parameter selection & 0.79 & 0.67 \\
        \bottomrule
    \end{tabular}
    \vspace{0.05in}
    \caption{Offline and task-time accuracy for robot learning.}
    \label{tab:retrieval}
\end{table}

In offline experiments, our algorithm demonstrates an accuracy of $94\%$ for the VLM object and task proposal and $79\%$ for the DINO parameter prediction. In online task-time experiments, our algorithm achieves an accuracy of over $83\%$ for the VLM object and task proposal and $79\%$ for the dino parameter prediction. Note the slight decrease in accuracies between offline to online demonstrations primarily stemmed from minor object shifts caused by the gripper during execution and small offsets introduced by the top-view camera that does not perfectly align in parallel to the table surface. 

What this means is that the user can skip object and skill selection $83\%$ of the time, and can skip parameter selection over $79\%$ of the time during a task. This significantly reduces the time and effort required from the user. Hence we see that with object and skill learning, the average time of human involvement is reduced by $12\%$ from $4.76$ to $4.18$ minutes.

\section{Discussion and Ethical Concerns}
The improved decoding framework of \algo 2.0 substantially increased the skill selection accuracy at task time. The transition from binary to 4-way classification with continuous, closed-loop control allows for real-time adjustments and re-selections, enhancing user control in 3D space. By integrating retrieval-based few-shot object and skill selection with one-shot skill parameter learning, NOIR 2.0 showed significant performance improvements. Through pre-trained vision-language models for state understanding and task retrieval, we reduced the amount of demonstration data required from 15 examples to 1 user demo and enabled more effective learning. With parameter prediction capability, we addressed challenges in object placement and generalizability. These changes contribute considerably to reducing overall task completion time and effort in performing everyday activities.

It is also important to acknowledge existing challenges in accurately decoding extracortical EEG data, as well as the subject-variability of demonstrable signals. User fatigue in concentrating for long periods of time on long-horizon tasks may affect the quality of extracted signals which translates to a decrease in system performance. The EMG-based interruption system using facial muscle tension continues to provide crucial and reliable safety features for \algo 2.0, allowing users to maintain control over the system. It is imperative to ensure the robustness and safety of end-to-end brain-robot systems like \algo, especially when they are operated in dynamic environments or human spaces.

\section{Conclusion}
In this paper, we introduced \algo 2.0, an enhanced version of the NOIR BRI. Through increased speed and accuracy when decoding brain activity and incorporating one-shot learning to predict human intent, \algo 2.0 facilitates more efficient human-robot collaboration. By leveraging the algorithm's ability to adapt to individual users and their intentions real-time, \algo 2.0 reduces cognitive burden and increases the system's adaptability This enables everyday users to manipulate a robot to perform various real-world tasks using brain signals with little to no training. We believe that \algo 2.0 holds a significant potential to augment human capabilities and to serve as a critical component of assistive technology. Its intuitive interface and natural interactive paradigm make it especially valuable for individuals who may require daily physical assistance. Integration of these technologies opens up new avenues for future research in collaborative human-robotics and the expansion of BRIs into other areas of human activity.

\bibliography{ref-neural,ref-robot}  %

\end{document}